\documentclass{article}
\usepackage{amsmath,graphicx,mlspconf}
\usepackage{xcolor}
\usepackage{algorithmic}
\usepackage{tikz}
\usepackage{amssymb}
\usepackage{float}
\usepackage{textcomp}
\usepackage[english]{babel}
\usepackage[nottoc]{tocbibind}
\usepackage{booktabs}
\usepackage{svg}
\usepackage{pgfplots}
\usepackage{setspace}
\usepackage{dsfont}
\usepackage{multicol}
\usepackage{blindtext}
\usepackage{subfigure}
\newcommand{\squeeze}{\vspace{-0.1in}}
\toappear{Accepted at 2023 IEEE International Workshop on Machine Learning for Signal Processing, Sept.\ 17--20, 2023, Rome, Italy}

\title{Robot Motion Prediction by Channel State Information}
\name{Rojin Zandi $^\dagger$, Hojjat Salehinejad $^\star$, Kian Behzad $^\dagger$, Elaheh Motamedi $^\dagger$,and Milad Siami $^\dagger$ \thanks{
This material is based upon work supported in part at Northeastern University by grants ONR N00014-21-1-2431, NSF 2121121, the U.S. Department of Homeland Security under Grant Award Number 22STESE00001-01-00, and by the Army Research Laboratory under Cooperative Agreement Number W911NF-22-2-0001. The views and conclusions contained in this document are solely those of the authors and should not be interpreted as representing the official policies, either expressed or implied, of the U.S. Department of Homeland Security, the Army Research Office, or the U.S. Government.}}

\address{$^\dagger$Department of Electrical \& Computer Engineering, Northeastern University, Boston, USA\\
$^\star$Kern Center for the Science of Health Care Delivery, Mayo Clinic, Rochester, MN, USA}

\begin{document}
\maketitle
\squeeze
\begin{abstract}
Autonomous robotic systems have gained a lot of attention, in recent years. However, accurate prediction of robot motion in indoor environments with limited visibility is challenging. While vision-based and light detection and ranging (LiDAR) sensors are commonly used for motion detection and localization of robotic arms, they are privacy-invasive and depend on a clear line-of-sight (LOS) for precise measurements. In cases where additional sensors are not available or LOS is not possible, these technologies may not be the best option. This paper proposes a novel method that employs channel state information (CSI) from WiFi signals affected by robotic arm motion.
We developed a convolutional neural network (CNN) model to classify four different activities of a Franka Emika robotic arm. The implemented method seeks to accurately predict robot motion even in scenarios in which the robot is obscured by obstacles, without relying on any attached or internal sensors.

\end{abstract}
\allowdisplaybreaks
\begin{figure*}
    \centering
    \includegraphics[width=0.89\textwidth, height=0.198\textwidth]{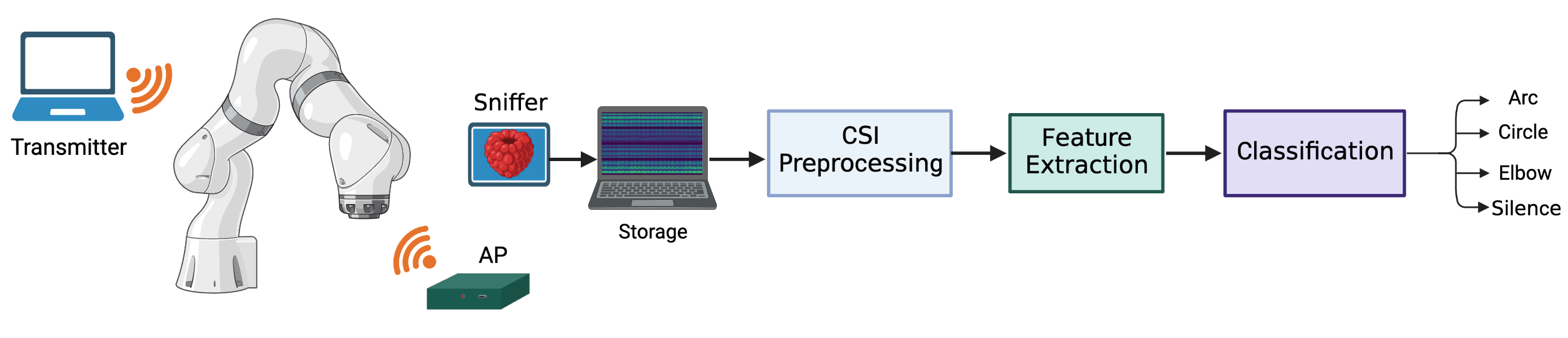}
    \caption{Flowchart of different steps of the proposed method, from CSI collection to classification.}
    \label{fig:setup}
\end{figure*}

\section{Introduction}

\label{sec:intro}
In recent years, the development of autonomous robotic systems alongside with the rapid advancements in artificial intelligence have gained significant attention due to their ability to operate in various environments without human intervention \cite{atkeson2015no}. Autonomous robots have a wide range of applications in different industries such as welding in manufacturing \cite{bischoff2010kuka}, bin-picking in agriculture \cite{nieuwenhuisen2013mobile}, and surgery \cite{lanfranco2004robotic}, where they can perform hazardous or redundant work with exceptional accuracy and precision. Although robots have experienced rapid development, accurately predicting robot motion remains one of the main challenges in robotics, especially in indoor environments with limited visibility \cite{correa2012mobile}. Light detection and ranging (LiDAR) sensors were utilized in visually challenged indoor environments \cite{li20152d, petrlik2021lidar}. However, accurate measurements require a clear line-of-sight (LOS); in circumstances where additional sensors are unavailable or LOS is not feasible, LiDAR may not be an appropriate solution. Another issue with vision-based methods is lack of privacy, particularly in surveillance systems.
While vision-based methods have the mentioned problems, it is possible to detect the human presence with WiFi signals instead of common sensors \cite{woyach2006sensorless}. To experimentally verify the result of human movement on the attenuated WiFi signal, authors in \cite{youssef2007challenges} applied simple model-based approaches such as moving average. To enhance human activity recognition results channel state information (CSI) can be utilized \cite{wang2014eyes}, which provides detailed information about the phase and amplitude of each subcarrier in multipath fading. This enables the detection of small movements that can alter a multipath environment \cite{adib20143d}. 

Despite WiFi systems being ubiquitously available and offering greater security compared to vision-based systems, they still present several challenges. First, using a single receiver or access point (AP) may lead to poor classification results depending on the indoor environment. Second, CSI is highly dependent of location and orientation, i.e., the collected data of a user with same activity is severely different in another room. Third, the complexity of human activity requires a sophisticated machine learning model, which may be prune to overfitting. These issues have been tackled by collecting a human motion dataset by one transmitter and six receivers \cite{10.1145/3307334.3326081}. Using multi receivers can be challenging because of synchronization process. The paper \cite{salehinejad2023joint} introduces a novel approach that predicts both the activity and orientation of subjects by employing multivariate minirocket feature extraction \cite{dempster2021minirocket} and a simple Ridge regression classifier. Using a standard CNN, RF signals have proven effective in detecting hand tremors. With their high sensitivity, RF signals enable precise analysis of subtle motion patterns \cite{9943307}.
WiFi pose estimation has also been implemented for mobile robots, by using the IEEE 802.11 wireless local-area network (WLAN) adapters on mobile robots to determine their position and orientation by measuring CSI of APs based on the Euclidean distance on a predefined radio map \cite{rohrig2008wlan}.

However, many arm robots are not equipped with a WiFi module and using these techniques are not helpful. So, our objective is to collect and study a dataset of robotic arms free of any specific equipment to be attached to for data collection and motion detection. Given the advantages of WiFi CSI data in privacy preserving and high performance in non-line-of-sight (NLOS) indoor environments, we collected CSI data of a Franka Emika robotic arm movement \cite{haddadin2022franka} with goal of classifying its activities in with LOS and NLOS indoor environments, and to best of our knowledge our study is the first of its kind.
\vspace{-0.4cm}
\squeeze
\section{Method}
\label{sec:method}
\begin{figure}[H]
    \centering
    \includegraphics[width=0.5 \textwidth]{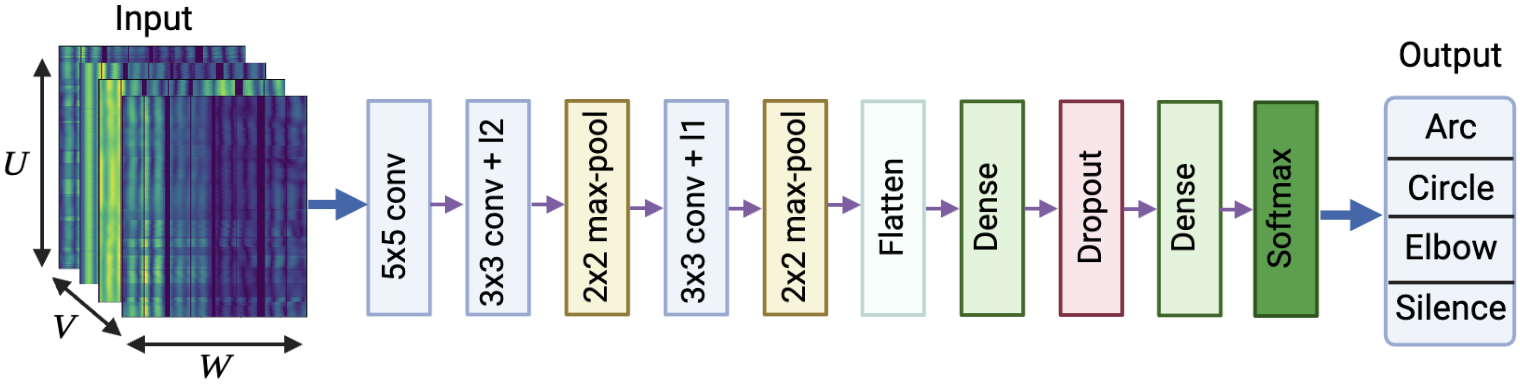}
    \caption{Architecture of designed CNN to classify the input CSI amplitudes.}
    \label{fig:cnn-arch}
\end{figure}

\squeeze
To study the similarity of CSI data caused by different types of robotic arm movements, one can apply convolutional neural networks (CNN). Considering the effectiveness of CNNs to capture temporal patterns in time-series, and their ability to learn local patterns and dependencies such as periodicity and trends of data in the feature extraction layers \cite{ismail2019deep}, applying them to CSI would be beneficial. Fig. \ref{fig:setup} illustrates the flowchart of collecting CSI data of the robotic arm motion, and then processing the data to classify four different motions. Fig. \ref{fig:cnn-arch} exhibits the taken approach to label the CSI data. First, the data is preprocessed by reshaping it to smaller samples, and then fed to the CNN model to extract the features and classify them to defined labels.

\subsection{Preprocessing}
\begin{figure}[H]
     \centering
     \includegraphics[width=0.46 \textwidth]{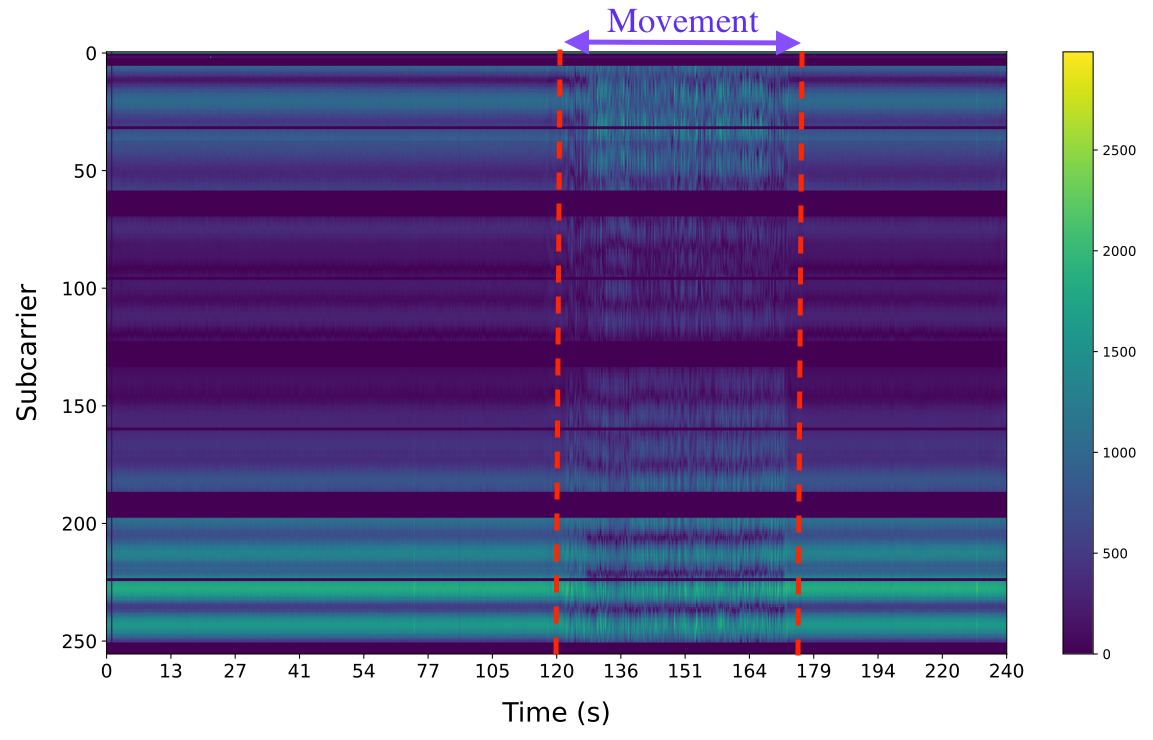}
     \caption{Amplitude of CSI data in time domain. The annotated area indicates the period of robotic arm movement, which has occurred between the $120^{th}$ and $176^{th}$ seconds.}
     \label{fig:csi-amp}
 \end{figure}

The collected CSI is presented as a matrix of complex numbers $\mathbf{H}\in \mathbb{C}^{N\times W}$ where $N$ and $W$ are number of transmitted packets and sub-carriers for each antenna, respectively. 
To feed the input data to our model, we extract the amplitude of the complex CSI values from $\mathbf{H}$ to train our model \cite{salehinejad2022litehar}. To attain the input samples we have reshaped matrix $\mathbf{H}$ to an input tensor $\mathcal{X} \in \mathbb{R}^{U \times W \times V}$, where each one of the frontal slices $\mathcal{X}_{::v} $ is an input sample to our CNN model. By mode-$v$ matrixization of $\mathcal{X}$, we reorder the input samples and achieve the original matrix $\mathbf{H}$.
 The advantage of using the amplitude of CSI data as a 2-D input sample is that it can capture the spatial and temporal characteristics of robot motion more precisely than using 1-D CSI data. This is because the 2-D amplitude data delivers a more precise picture of the wireless signal, enabling the CNN to capture the dynamics of robot movement, with higher accuracy.   \\
 Fig. \ref{fig:csi-amp} presents a 2D colour map of the amplitude of CSI for 256 sub-carriers over time, and the disturbance resulted from environmental movement, which has occurred between the $120^{th}$ and $176^{th}$ seconds, as indicated by the arrow highlighting the robotic arm movement's duration. \\
 \vspace{-0.5cm}
\vspace{-0.53cm}


        
 
\begin{figure*}[t]
\centering
\subfigure[Scenario 1]{\includegraphics[width=4.35cm, height=3.5cm]{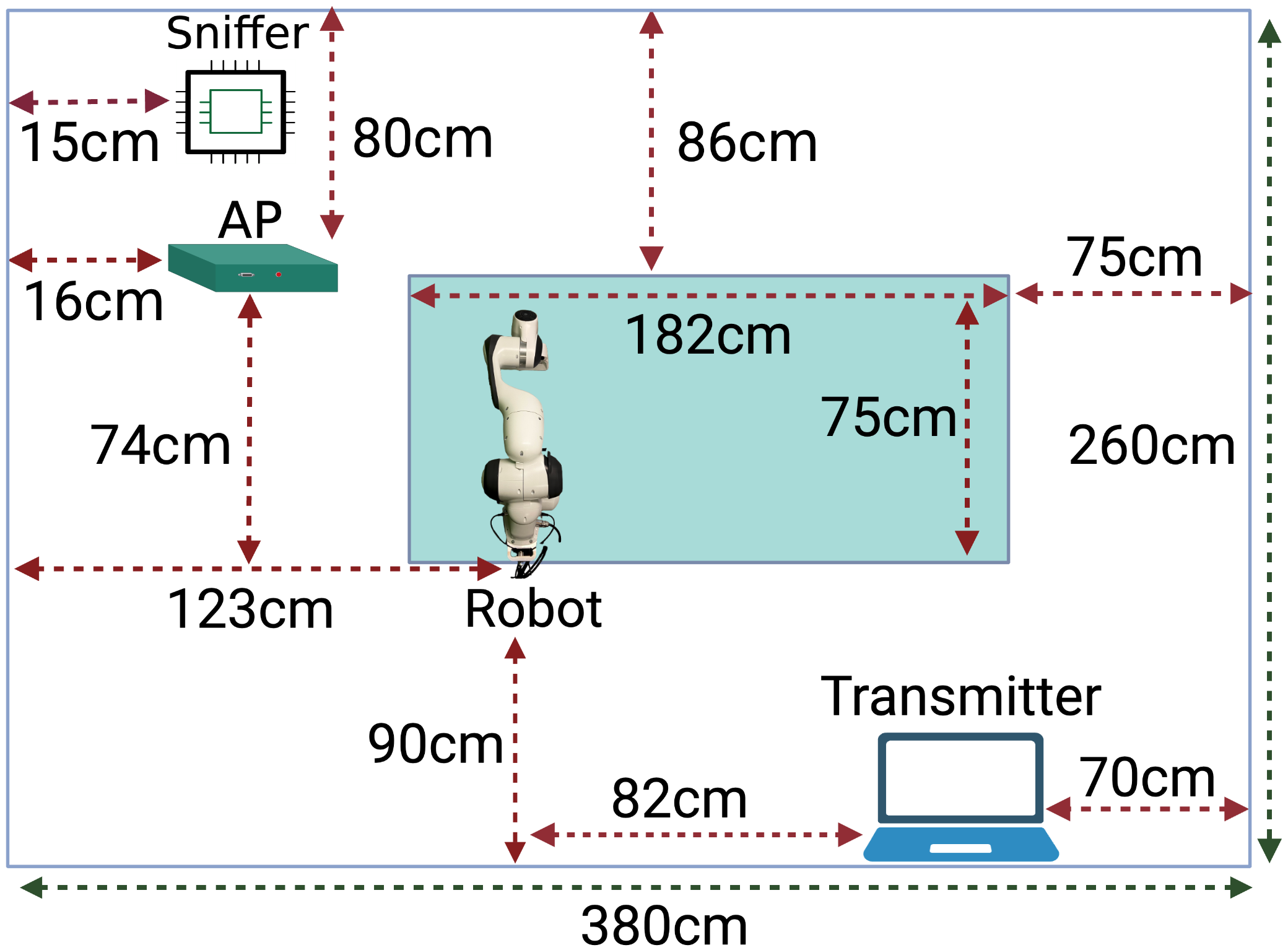}}
\subfigure[Scenario 2]{\includegraphics[width=4.35cm, height=3.5cm]{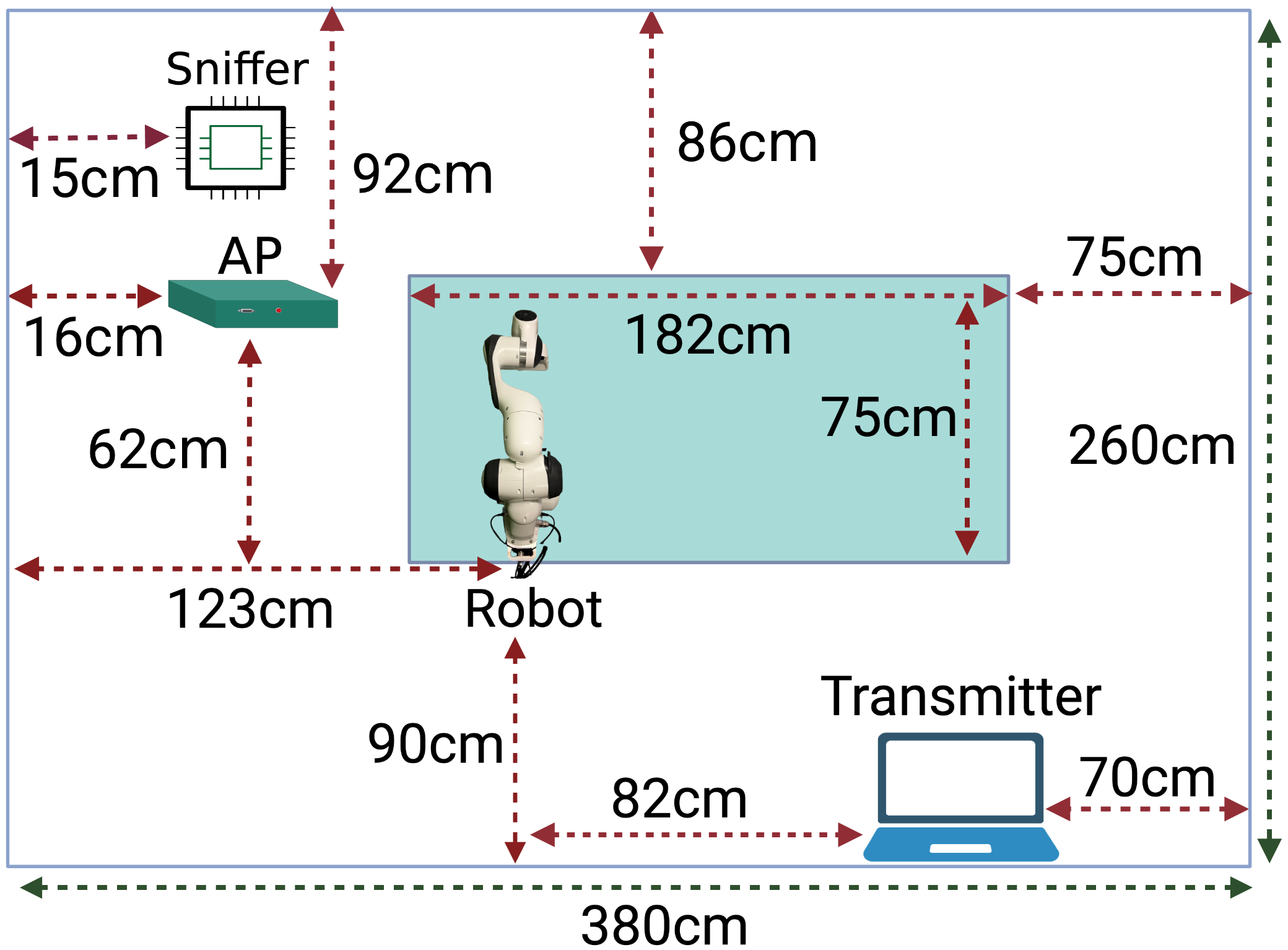}}
\subfigure[Scenario 3]{\includegraphics[width=4.35cm, height=3.5cm]{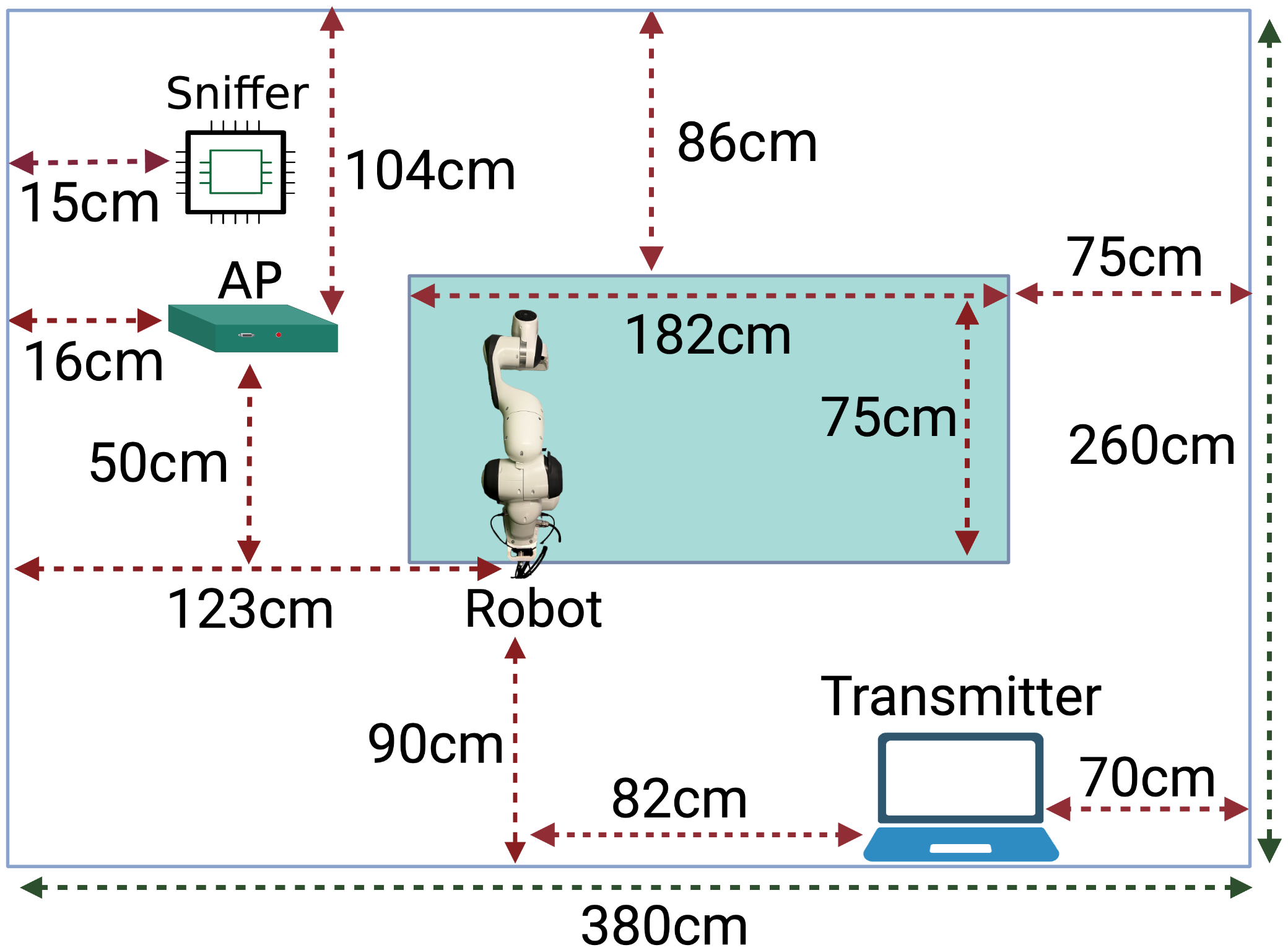}}
\subfigure[Scenario 4]{\includegraphics[width=4.35cm, height=3.5cm]{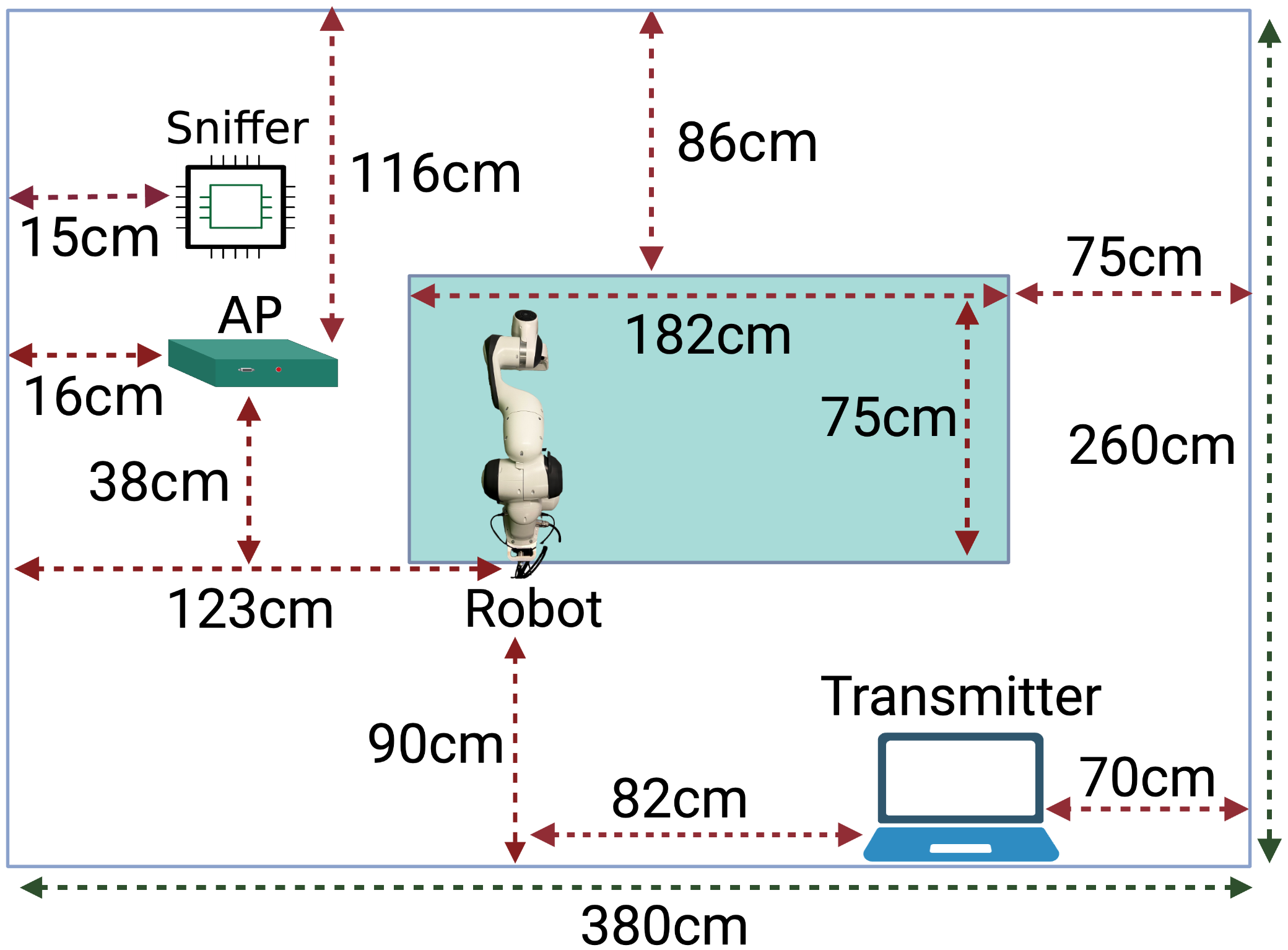}}
\caption{Different data collection scenarios. In each scenario, the receiver and sniffer have moved 12 cm horizontally.}
\label{fig:scanrios}
\end{figure*}
\subsection{Classification}
In this section, we describe the structure of our implemented CNN model. As depicted in Fig. \ref{fig:cnn-arch}, the model has been developed in three 2-D convolutional, two max-pool to extract the features, and finally two dense layers with rectified linear unit (ReLU) and softmax activation functions for classifying the data. In order to decrease the overfitting, we have applied Lasso and Ridge regression for learning the kernels using $l1$ and $l2$ regularization and also dropout method in between dense layers \cite{wager2013dropout}. 
\begin{figure}[H]
\centering
\subfigure[Arc]{\includegraphics[width=0.216 \textwidth]{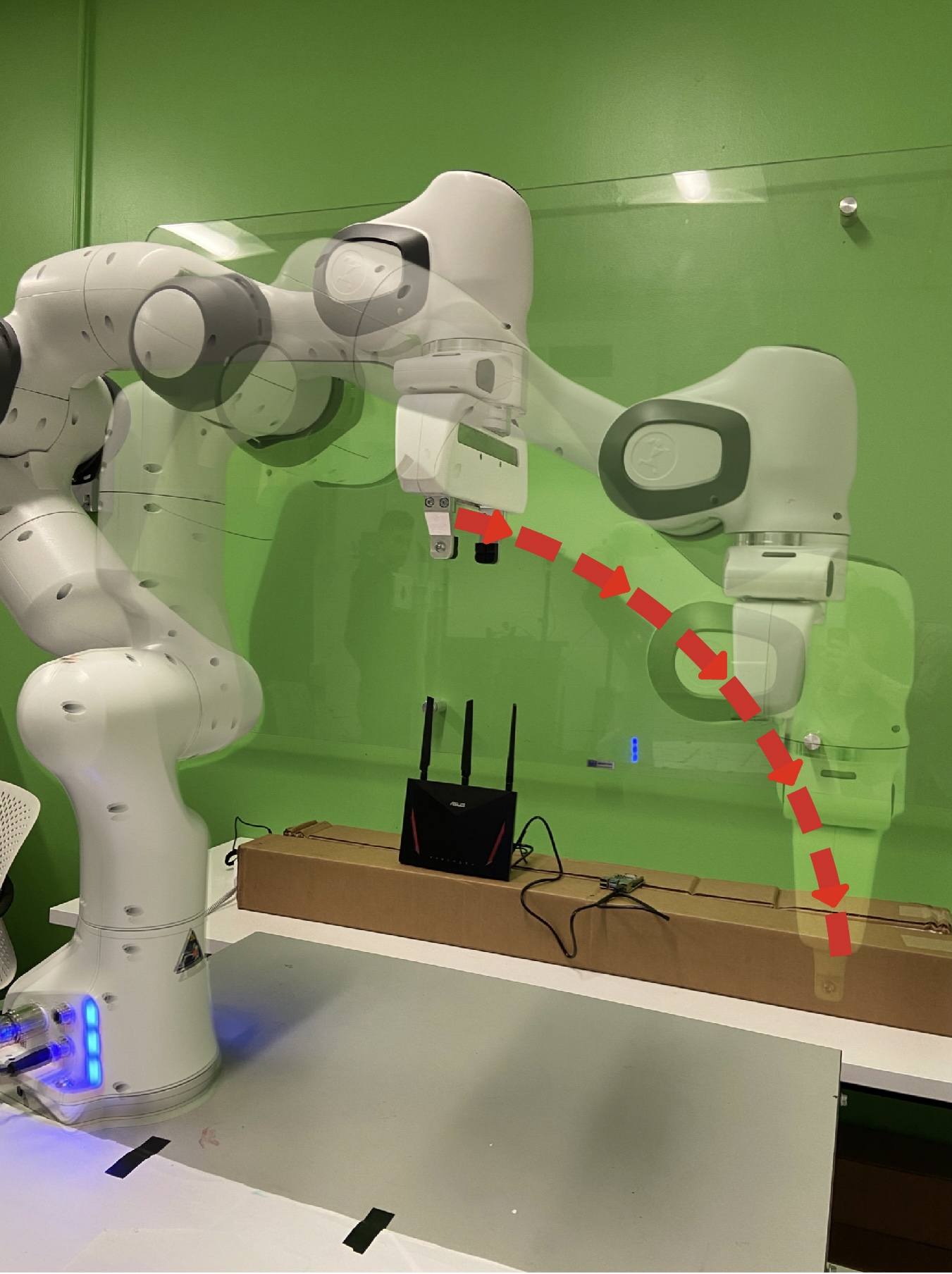}}
\subfigure[Elbow]{\includegraphics[width=0.216 \textwidth]{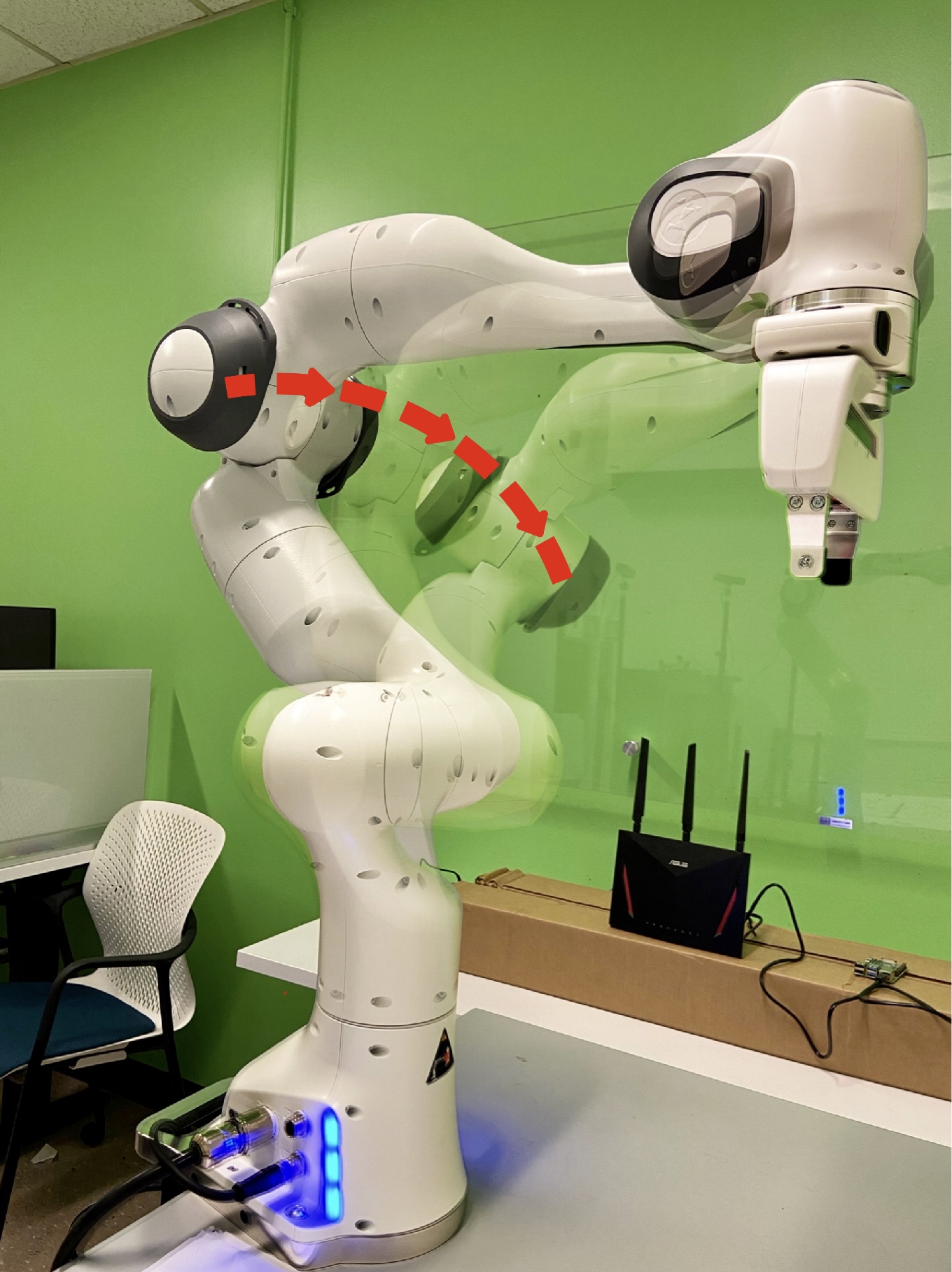}}
\subfigure[Circle]{\includegraphics[width=0.216 \textwidth]{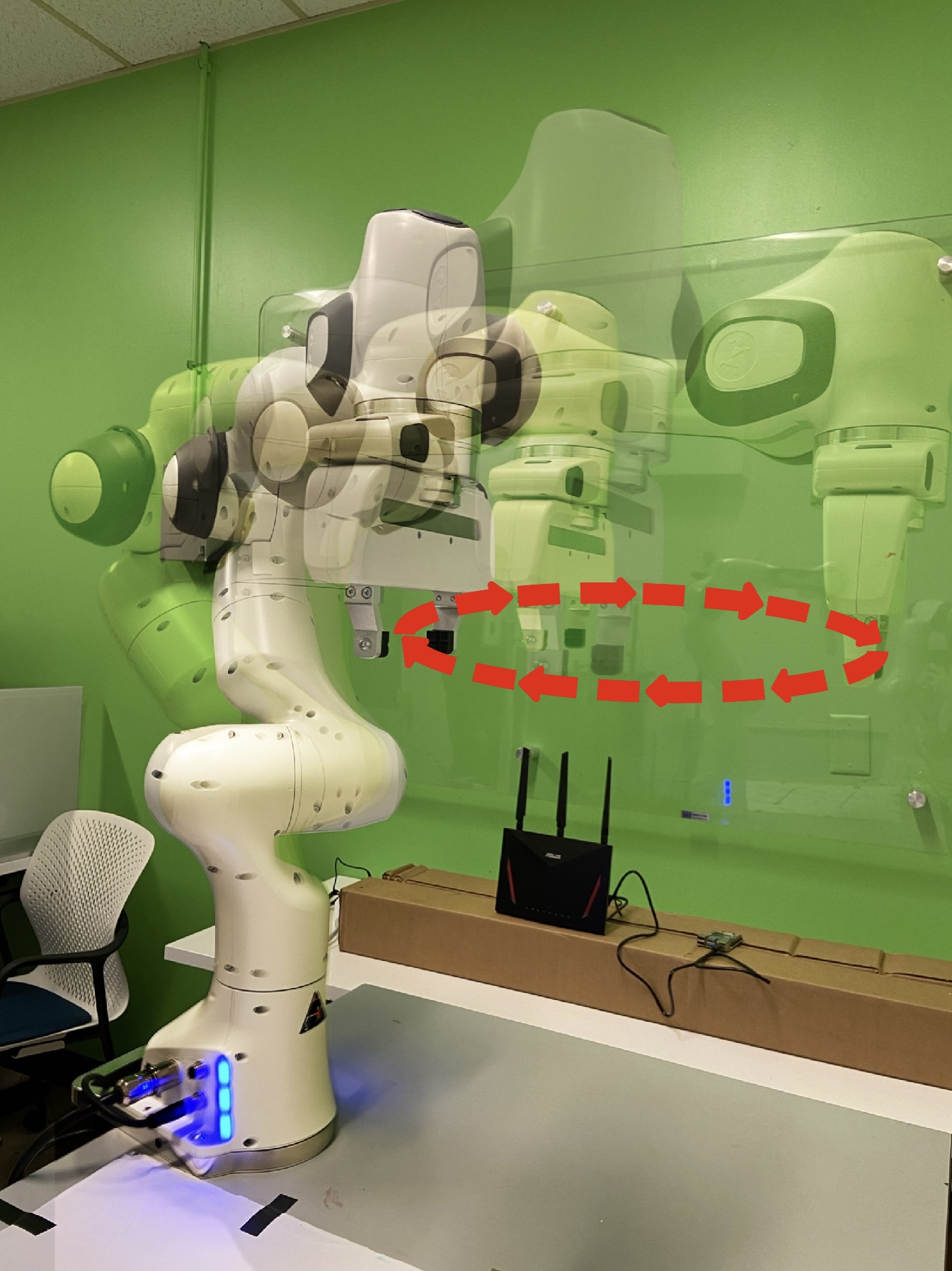}}
\subfigure[Silence]{\includegraphics[width=0.216 \textwidth]{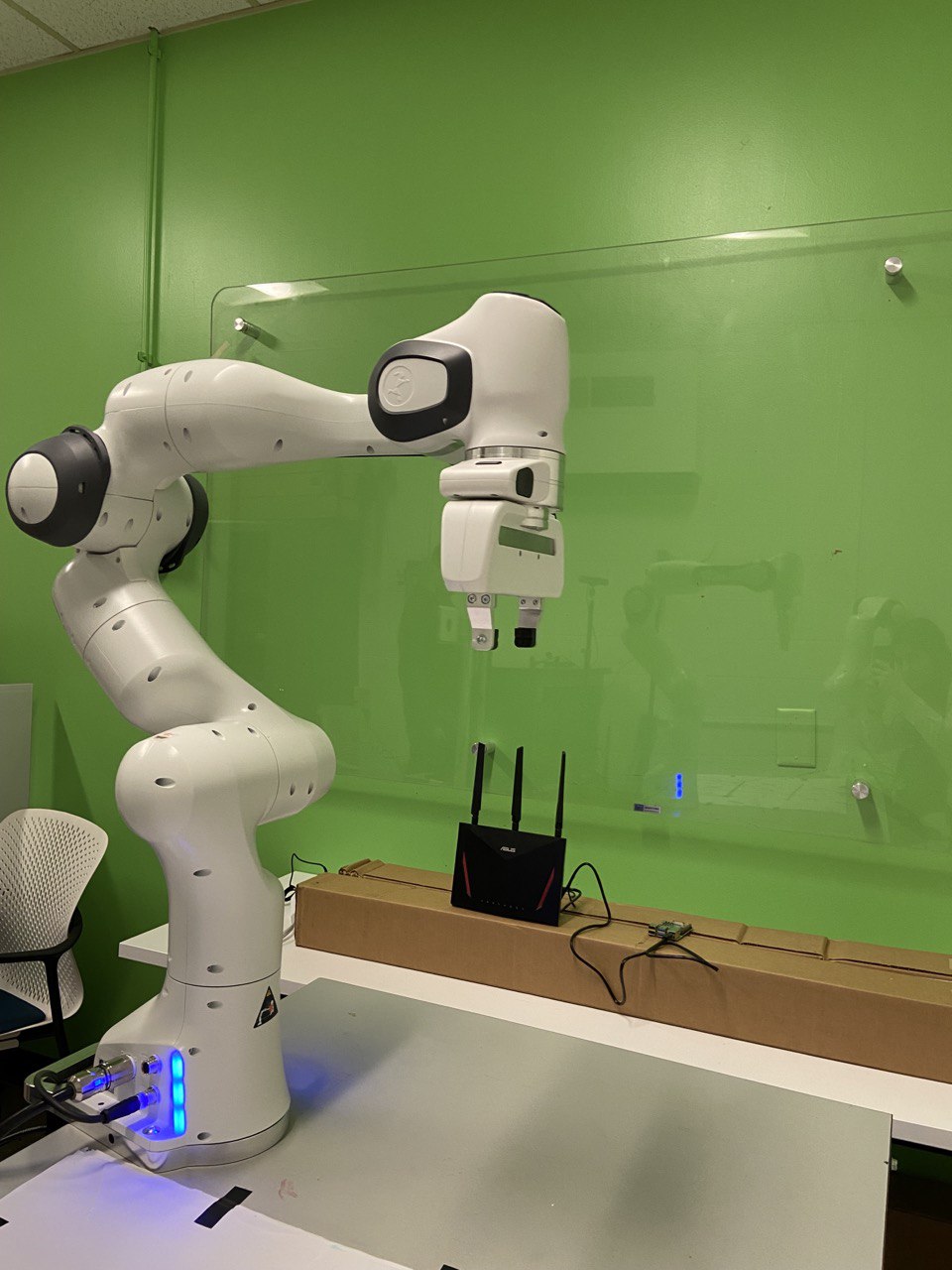}}
\caption{Illustration of the four actions performed by the Franka Emika arm in the experiments: (a) \textit{Arc}, (b) \textit{Elbow}, (c) \textit{Circle} in XY axis, and (d) \textit{Silence}. Each action was executed for over 5 minutes at a 30 Hz frequency, generating 10,000 packets of data.}
\label{fig:actions}
\end{figure}
\vspace{-0.2cm}
\section{Experiments}\label{sec:exp}
\subsection{Data Collection}
We collected data of 3 different actions of a Franka Emika arm and one class of no movement, which are presented in Fig. \ref{fig:actions}. In the experiments the arm performed: (a) \textit{Arc}, (b) \textit{Elbow}, (c) \textit{Circle} in XY axis and (d) \textit{Silence}, and the red dashed arrows portray the trajectory of joints movements. Each action has been repeated for more than 5 minutes with 30 Hz frequency, and 10000 packets have been received by the router. The bandwidth is set on 80 MHz which results in using 256 sub-carriers. For each sample, we used 300 consecutive CSI packets from the environment, resulting in a matrix of size 300$\times$256 and prior to training, we preprocess the data by removing 8 pilot and 14 unused subcarriers \cite{10.5555/2563615}, so the each input sample is reshaped to 300$\times$234. To ensure sufficient diversity in our dataset, we collected 100 samples for each class, resulting in a final CSI data of size 400$\times$300$\times$234 for four classes of action.
As a transmitter, we used a laptop with IEEE 802.11ac standard to send the packets to an ASUS RT-AC86U router which is set on 5 GHz and on channel 36 with similar IEEE standard.  

To gather the receiving CSI data, we utilized a Raspberry Pi B4 configured as a sniffer, as illustrated in Fig. \ref{fig:setup}, equipped with a bcm43455c0 WiFi chip. The Raspberry Pi B4 is capable of computing and transmitting CSI information over the local network and to a computer for further analysis. After collecting the data, CSI has been monitored and stored in a laptop, and the data preprocessing and classification has been performed in the same laptop.
\vspace{-0.6mm}
\subsection{Training Setup}
\begin{figure}[H]
    \centering
    \includegraphics[width=0.44 \textwidth]{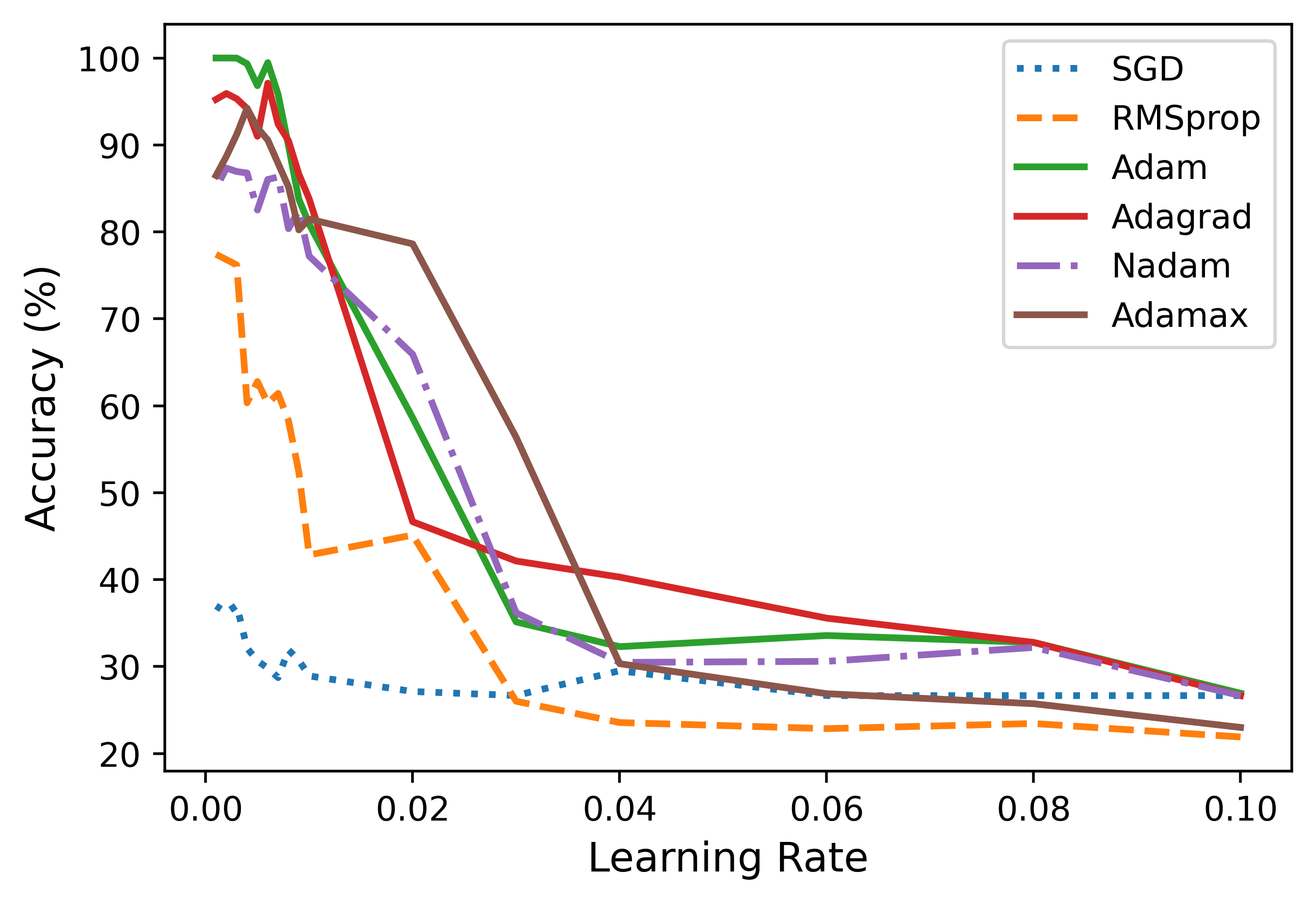}
    \caption{CNN model accuracy using different optimization methods. Adam and Adagrad can attain higher accuracy values.}
    \label{fig:opt-met}
\end{figure}

\vspace{-0.3mm}
To train the CNN model, we use batch size of 16 to control the time of convergence and amount of overfitting \cite{ioffe2015batch} during 50 epochs which are also controlled by early stopping with 6 epochs patience monitoring the accuracy. The applied loss function and optimizer are categorical cross entropy and Adam optimizer \cite{kingma2014adam} with $0.001$ learning rate, respectively. 
To choose the optimizer and learning rate, we have implemented a grid search with 0.01 step size, and studied the range of 0.001 to 0.1 in stochastic gradient decent (SGD), RMSprop, Adam, Adagrad, Nadam, and Adamax and the results are shown in Fig. \ref{fig:opt-met}.
\vspace{-0.8cm}
\subsection{Results Analysis}
To evaluate the performance of the classification model, we have designed three different case studies. In each case we have computed precision, recall, F1-score, and accuracy for various scenarios of data collection, which are illustrated in Fig. \ref{fig:scanrios}.
\subsubsection{5-fold Cross-Validated data with Line-of-Sight}\label{sec:no-obs}
\begin{figure}[H]
\centering
\subfigure[Scenario 1]{\includegraphics[width=0.22 \textwidth]{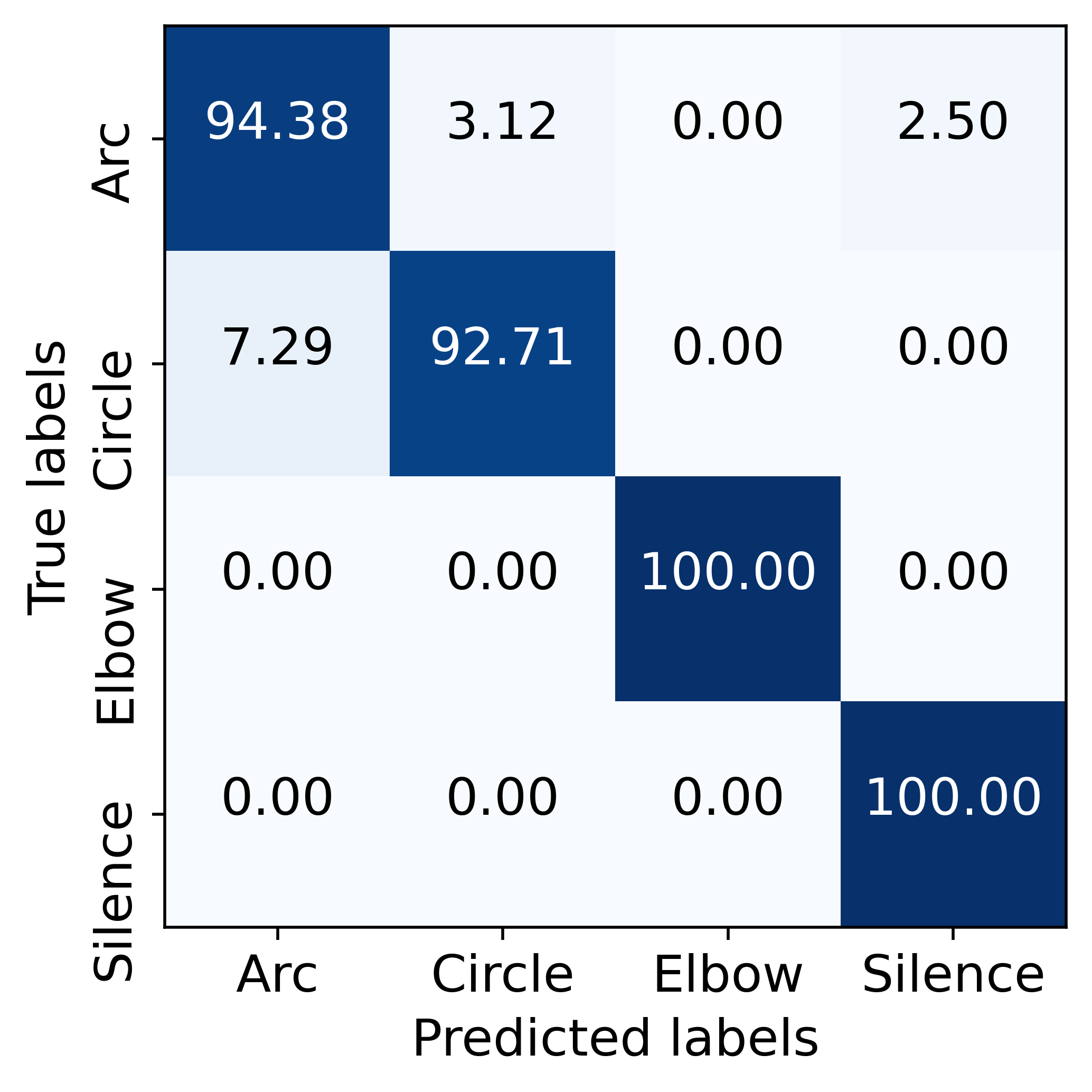}}
\subfigure[Scenario 2]{\includegraphics[width=0.22 \textwidth]{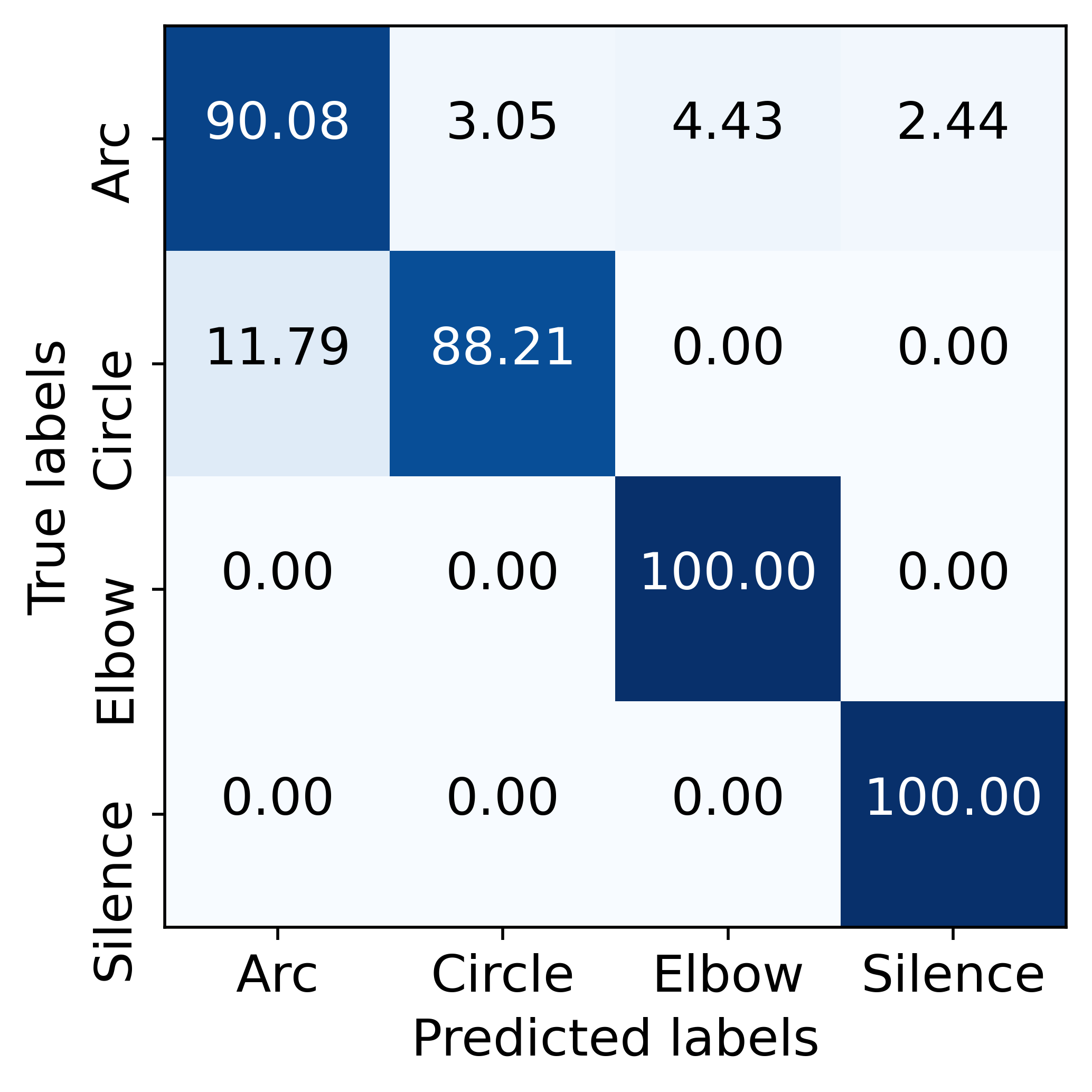}}
\subfigure[Scenario 3]{\includegraphics[width=0.22 \textwidth]{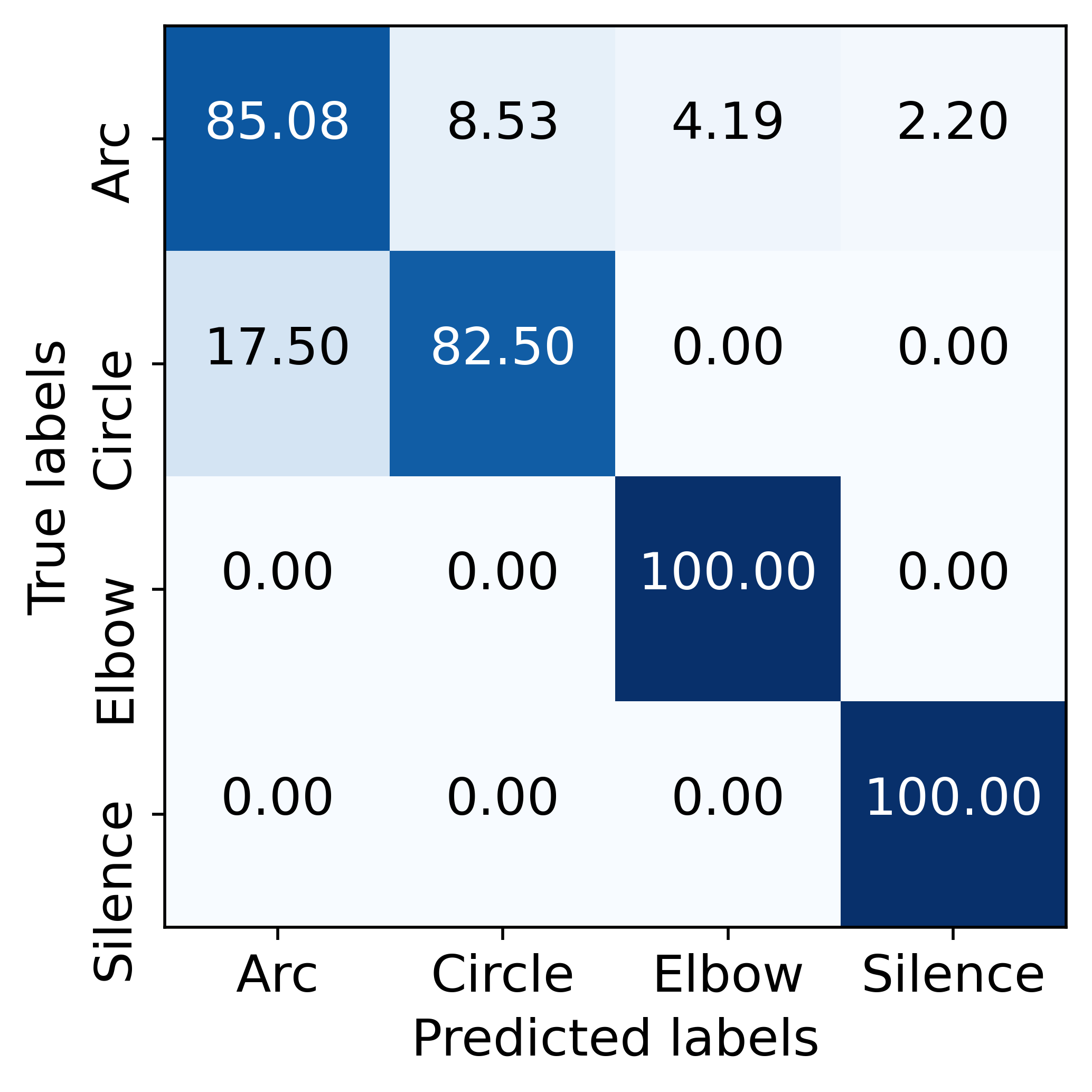}}
\subfigure[Scenario 4]{\includegraphics[width=0.22 \textwidth]{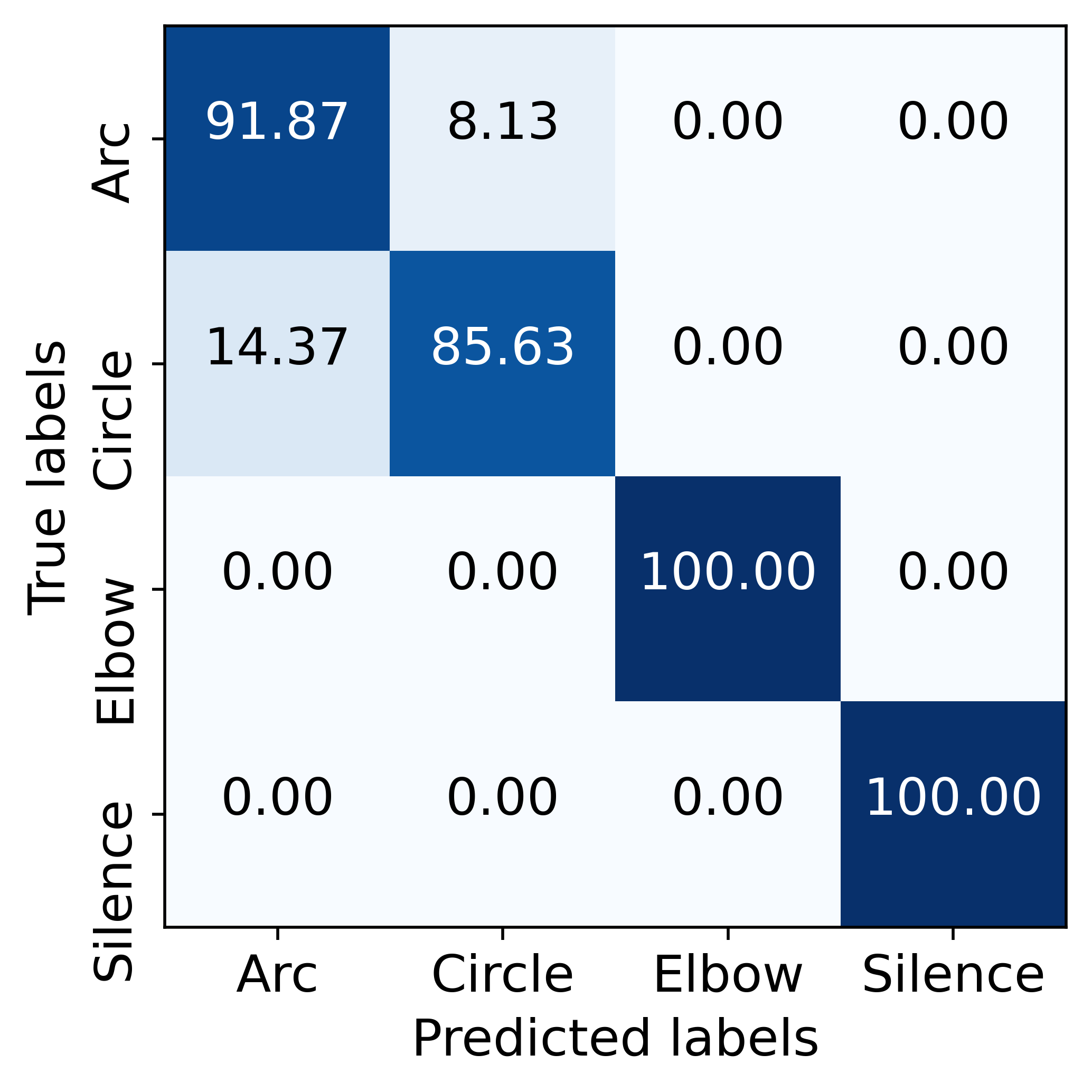}}
\caption{After 5-fold cross-validation, we obtained the average confusion matrix for various data collection scenarios, presented in percentage. The circle and arc classes showed lower accuracy rates, likely due to the similar motion patterns of the Franka Emika robotic arm involved.}
\label{fig:CM-scanrios}
\end{figure}

As mentioned in Section \ref{sec:intro}, changes in indoor environment such as location and orientation can affect the resulted CSI, so we study the data attained from four different scenarios, which are illustrated in Fig. \ref{fig:scanrios}. In each case, the transmitter and robot are static and the only change in the room is the location of the receiver for 12 cm and we collect the data of the four mentioned classes. The model has been trained and tested four different times for each scenario, and the train and test data have been split with 20\% ratio. The results shown in Fig. \ref{fig:CM-scanrios} are the average of confusion matrix of 5-fold cross-validation over each scenario. The average precision, recall, F1-score and their corresponding standard deviation after cross-validation has been presented in  Table \ref{tab:sep-sc}. The test and train data are balanced so we anticipate to observe equal accuracy and recall. 

\begin{table}[H]
\footnotesize
    \centering
    \caption{Average and STD of precision, recall, F1-score and accuracy of each scenario after 5-fold cross-validation in percentage.}
    \vspace{0.2cm}
    \begin{tabular}{|c|c|c|c|}
    \hline
    Scenario & Precision $(\%)$ & Recall $(\%)$& F1-Score $(\%)$\\
    \hline 1 & 97.20$\pm$ 0.28 & 96.34$\pm$0.58 & 96.64$\pm$0.22 \\
    \hline 2 & 95.11$\pm$1.22 & 93.65$\pm$1.96 & 93.69$\pm$1.92 \\
    \hline 3 & 93.63$\pm$2.54 & 93.02$\pm$1.96 & 93.01$\pm$2.29 \\
    \hline 4 & 95.12$\pm$1.73 & 94.46$\pm$1.85 & 94.42$\pm$1.75 \\
    \hline
    \end{tabular}
    \label{tab:sep-sc}
\end{table}

\vspace{-0.3cm}
Evidently, detecting the silence and elbow motion are more facile, in comparison with the circle and arc motion, based on Fig. \ref{fig:CM-scanrios}. This issue arises from a couple of causes;  during the elbow motion, the position of the end effector is fixed and only the intermediate joints move, whereas during the arc and circle actions the end effector and last link perform the majority of the motion. Hence, these two actions primarily obstruct common channels in space, which results in having more similar disturbances in CSI data distribution. Another reason for this issue is the length of these movements. To have a full cycle of one circle which takes 15 seconds, we must have 450 packets in each sample, \textit{i.e.} the sample size must be increased to at least 450$\times$234. 
\squeeze
\subsubsection{Non-Line-of-Sight}
\begin{figure}[H]
    \centering
    \includegraphics[width=7.2cm, height=4.6cm]{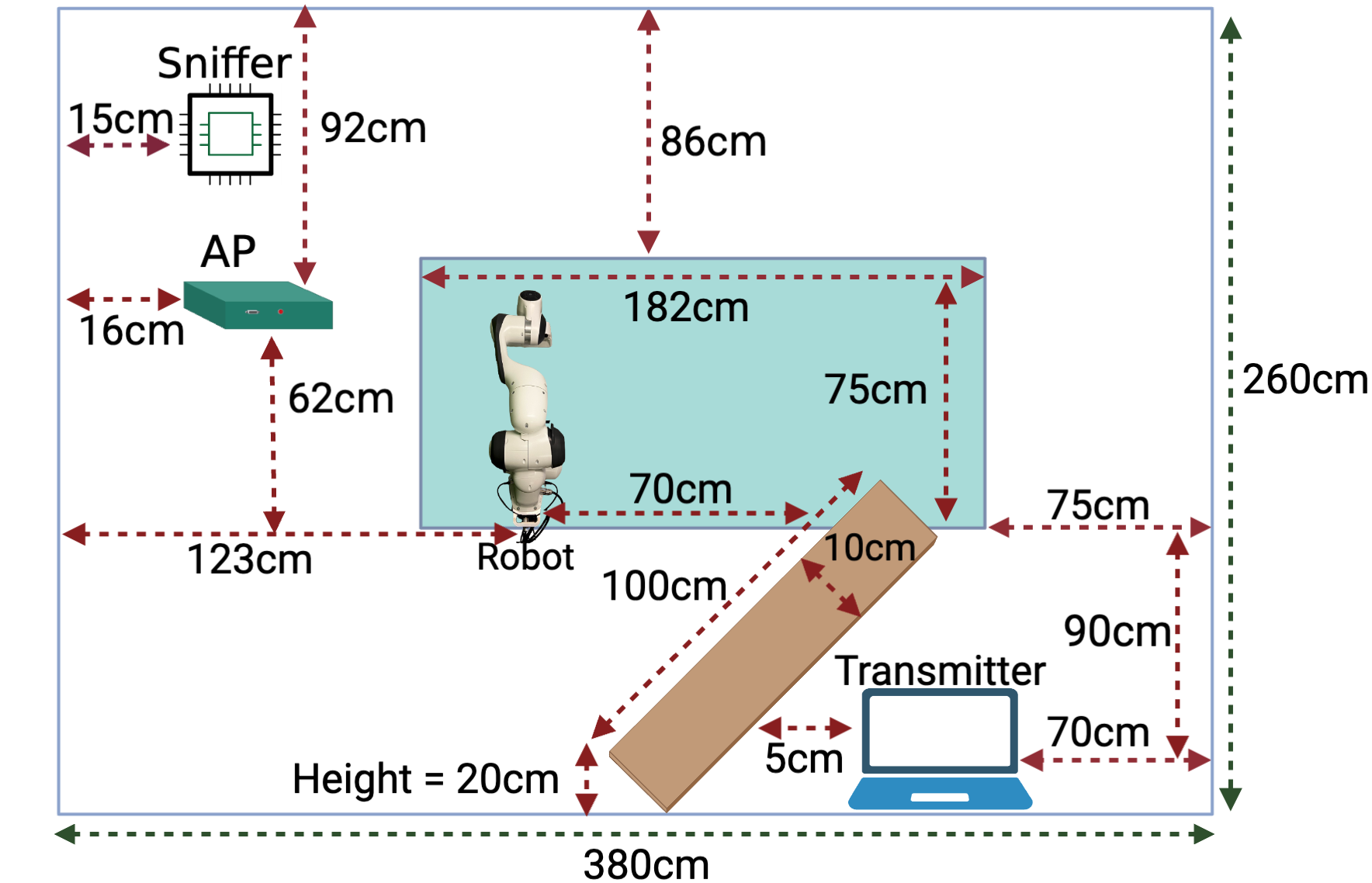}
    \caption{Floor plan with an obstacle in front of the transmitter.}
    \label{fig:obstacle}
\end{figure}

To study the effect of having an obstruction in between receiver and transmitter, during the data collection process, a filled card-box with $10\times20\times100$ cm dimensions has been used as an obstacle in front of the transmitter in scenario 2, as illustrated on Fig. \ref{fig:obstacle}. We have repeated data collection process for all four classes of activity while having the card-box as an obstacle, so we can compare the performance of the CNN model in the LOS and NLOS indoor environment.

\begin{figure}[H]
    \centering
    \subfigure[Classification metrics]{\includegraphics[width=4.1cm, height=3.4 cm]{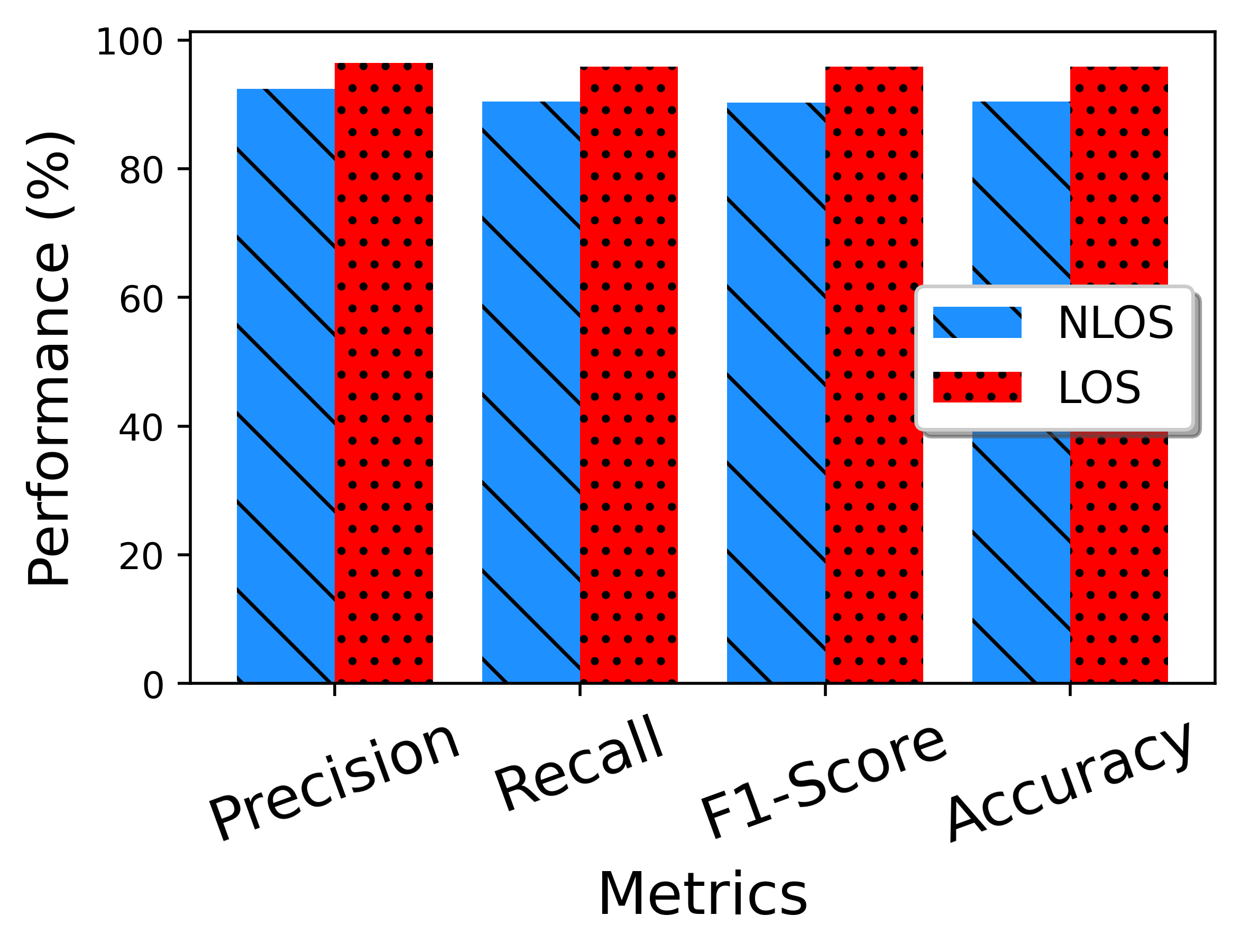} \label{fig:obs-met}}
    \subfigure[Accuracy per activity class]{\includegraphics[width=4.1cm , height=3.4cm]{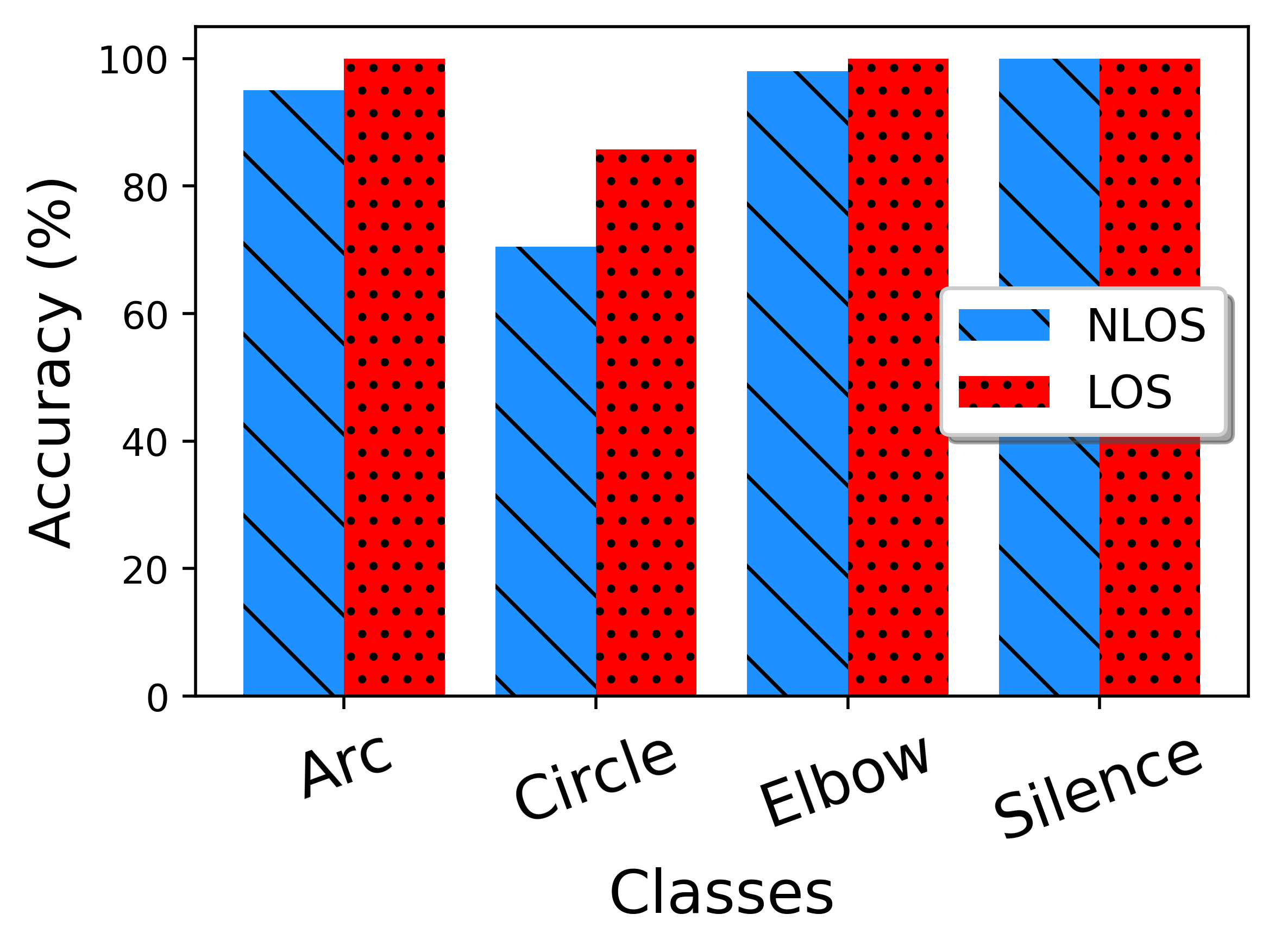} \label{fig:obs-acc}}
    \caption{Classification metrics and accuracy of each activity using data collected in LOS and NLOS indoor environments.}
    \label{fig:obs}
\end{figure}

Fig. \ref{fig:obs-met} conveys the classification performance metrics of the CNN model in case of having an obstacle during data collection with the case of having the same indoor environment without any obstacle in between the transmitter and the receiver except the robotic arm. While there is a decrease in model performance, the results demonstrate the ability of CSI data to distinguish between different types of motion even with an obstacle present. This is in contrast to vision-based methods, which struggle to classify activities when there is an obstruction in front of the camera. The classification accuracy of each type of activity is also shown in Fig. \ref{fig:obs-acc}, indicating that the model performs similarly in both with LOS and NLOS indoor environments. Additionally, detecting circular movement is more challenging than the other activities in both indoor environments.
\vspace{-0.2cm}
\vspace{-1.2cm}
\subsubsection{Leave-One-Scenario-Out with Line-of-Sight}
To establish a more resilient model in case of alternative receiver positions, we have cross-validated our model with the collected datasets from previous scenarios, \textit{i.e.}, the model has been trained on data from three scenarios and we left one scenario out to test the trained model on it. 
Table \ref{tab:cros-val} presents the average classification metrics and their corresponding standard deviations for the cross-validated model in each class of motion.

\begin{table}[H]
\footnotesize
    \centering
    \caption{Average and STD of precision, recall and F1-score of leave-one-scenario-out for different classes of action.}
    \vspace{0.2cm}
    \begin{tabular}{|c|c|c|c|}
    \hline
    Class & Precision $(\%)$& Recall $(\%)$& F1-Score $(\%)$\\
    \hline Arc & 83.01$\pm$8.00 & 79.64$\pm$10.74 & 81.16$\pm$9.00 \\
    \hline Circle & 77.31$\pm$11.97 & 75.66$\pm$8.61 & 75.21$\pm$2.46\\
    \hline Elbow & 98.43$\pm$1.70 & 100.00$\pm$0.00 & 99.19$\pm$0.40\\
    \hline Silence & 93.64$\pm$6.20 & 93.32$\pm$5.50 & 92.62$\pm$6.66\\
    \hline
    \end{tabular}
    \label{tab:cros-val}
\end{table}
\vspace{-0.2cm}
As discussed in Section \ref{sec:no-obs}, classifying the arc and circle motion are more complex, due to their similarity in the end-effector movement, and longer length of one full cycle of circle motion. Moreover, after cross-validation, we observe a drop in accuracy of the silent class, while elbow motion is still classified accurately. The repetitive motion of the elbow produces a periodic fluctuation in the amplitude of the received WiFi signal. This can aid the model in classification compared to the class of silence where there is no motion and the data is solely based on signal reflection. In this scenario, the CSI is more dependent on the stochastic noisy reflections from the indoor environment, making it more challenging for the model to identify a meaningful pattern and accurately predict the class because of having higher stochasticity.
\begin{figure}[H]
    \centering
    \includegraphics[width=0.28 \textwidth]{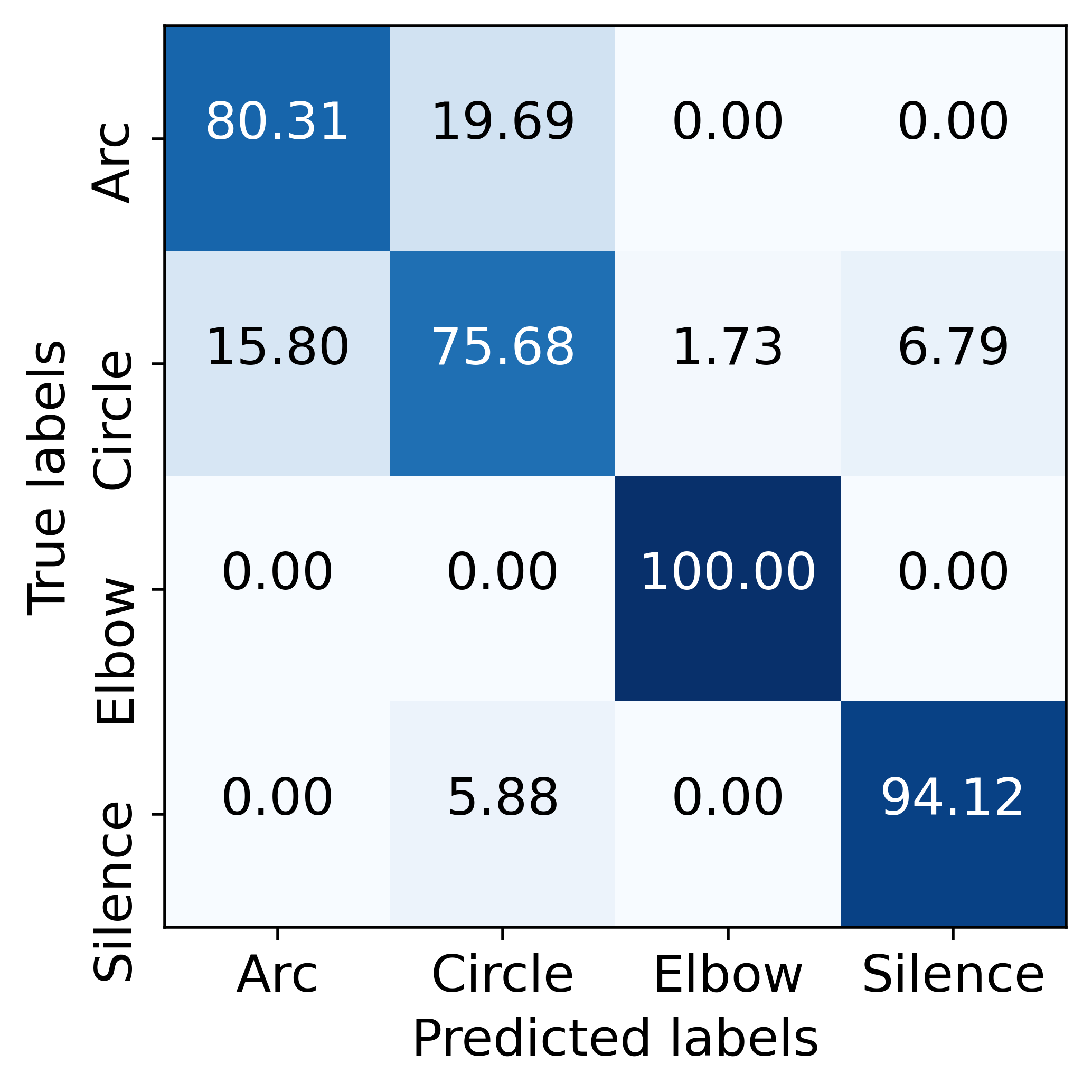}
    \caption{Average of confusion matrices of the leave-one-scenario-out in percentage.}
    \label{fig:CM-Cv}
\end{figure}
\section{Conclusions}
In conclusion, this work presents a novel method for robotic arm motion classification that makes use of CSI derived from WiFi signals. This method is effective even in scenarios in which the robotic arm is hindered by impediments. Given that the implemented CNN model was able to achieve accurate prediction of four different types of robotic arm motion, it demonstrated the potential of utilizing WiFi signals for robotic arm motion detection in indoor environments where traditional vision-based and LiDAR sensors may not be feasible. These types of indoor environments do not require any additional sensing and can make use of WiFi signals for robotic arm motion detection. This technique shows promise for use in a variety of domains, including robotics, surveillance, and the interaction between humans and robotic arms, and as a future research, one can study the effect of increasing APs in NLOS indoor environments.

\bibliographystyle{IEEEbib}
\ninept
\bibliography{strings,refs}
\end{document}